\documentclass[11pt]{article}

\usepackage[final]{acl}

\usepackage{times}
\usepackage{latexsym}

\usepackage[T1]{fontenc}

\usepackage[utf8]{inputenc}

\usepackage{microtype}

\usepackage{inconsolata}

\usepackage{graphicx}
\usepackage{booktabs}
\usepackage{amsmath}
\usepackage{amsfonts}
\usepackage{amssymb}
\usepackage[table]{xcolor}
\usepackage{xcolor}
\usepackage{tcolorbox}
\usepackage{multirow}
\usepackage{tabularx}
\usepackage{colortbl}
\usepackage{subcaption}
\tcbuselibrary{skins, breakable}
\usepackage{placeins}

\definecolor{lightgray}{gray}{0.9}
\definecolor{darkblue}{RGB}{0, 51, 102}

\newtcolorbox{examplebox}[1]{%
  enhanced,
  colback=lightgray!20,
  colframe=darkblue,
  coltitle=white,
  fonttitle=\bfseries,
  title=#1,
  boxrule=0.5mm,
  sharp corners,
  breakable,
  before upper={\small\obeylines\setlength{\parindent}{0pt}\setlength{\parskip}{0pt}}, %
  after upper={\par}
}

\title{Enhancing Linguistic Competence of Language Models through Pre-training with Language Learning Tasks}

\author{
Atsuki Yamaguchi \quad Maggie Mi \quad Nikolaos Aletras\\ 
School of Computer Science, University of Sheffield, United Kingdom\\
\texttt{\{ayamaguchi1,zmi1,n.aletras\}@sheffield.ac.uk}
}

\begin{document}

\maketitle
\begin{abstract}
Language models (LMs) are pre-trained on raw text datasets to generate text sequences token-by-token.
While this approach facilitates the learning of world knowledge and reasoning, it does not explicitly optimize for linguistic competence.
To bridge this gap, we propose \textbf{L2T}, a pre-training framework integrating \textbf{L}anguage \textbf{L}earning \textbf{T}asks alongside standard next-token prediction.
Inspired by human language acquisition, L2T transforms raw text into structured input-output pairs to provide explicit linguistic stimulation.
Pre-training LMs on a mixture of raw text and L2T data not only improves overall performance on linguistic competence benchmarks but accelerates its acquisition, while maintaining competitive performance on general reasoning tasks.\footnote{Our code and models are available via \url{https://github.com/gucci-j/l2t}.}\looseness=-1
\end{abstract}

\section{Introduction}
Language models (LMs) are pre-trained using causal language modeling (CLM), a next-token prediction objective~\cite{radfordimproving}.
While CLM pre-training equips LMs with world knowledge and general  capabilities~\cite{NEURIPS2020_1457c0d6,NEURIPS2022_8bb0d291}, 
it does not optimize for linguistic competence, i.e., the capacity to comprehend and interpret diverse linguistic phenomena~\cite{chomsky1965,waldis-etal-2024-holmes}.

Consequently, LMs often behave as stochastic parrots~\cite{10.1145/3442188.3445922}.
They mimic surface-level patterns without grasping the underlying linguistic scaffolding~\cite{chang-bergen-2024-language,lopezotal2025linguisticinterpretabilitytransformerbasedlanguage}.
This phenomenon mirrors rote learning in humans~\cite{PLUNKETT199321}, where learners reproduce patterns without understanding the generative rules~\cite{de9b4dc1-53a3-3120-90d0-e16fe9325393,https://doi.org/10.1111/0023-8333.00045}.

\begin{figure}[!t]
\centering
\includegraphics[width=0.98\columnwidth]{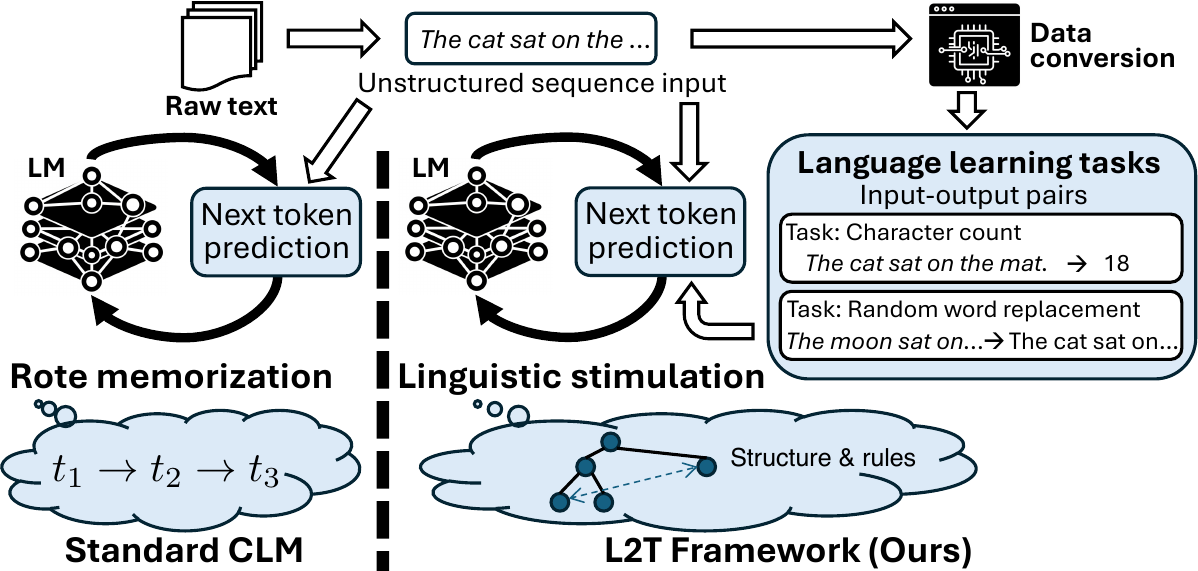}
\caption{
\textbf{L2T} vs. standard CLM over raw text.
}
\label{fig:motivation}
\end{figure}

We hypothesize that pre-training on \textbf{language learning tasks}, which require processing beyond raw sequence reconstruction, can stimulate the development of linguistic competence.
Specifically, we target data-efficient acquisition of morphological, syntactic, and semantic knowledge.
Inspired by human language acquisition, this approach encourages the development of structured representations that go beyond surface-level co-occurrence~\cite{AlishahiAfra2011CMoH,PERFORS2011306,doi:10.1073/pnas.1320525111}.
Evidence suggests that both humans and neural networks benefit from structured linguistic input during learning~\cite{ELMAN199371,GalkeLukas2024Dnna,hu-etal-2025-circuits}.\footnote{Appendix \ref{appendix:related_work} provides a detailed review of related work.}

We propose \textbf{L2T}, a pre-training framework integrating \textbf{L}anguage \textbf{L}earning \textbf{T}asks alongside standard CLM (Figure \ref{fig:motivation}).
Unlike instruction tuning requiring external supervision, L2T induces structure directly from raw text.
By converting text into structured input-output pairs, it enables the model to learn dependencies and how to restructure information, providing explicit linguistic stimulation absent in standard CLM.
We evaluate L2T by pre-training LMs on a mixture of L2T data and raw text at scales of 500M and 1B parameters.
Our main \textbf{contributions} are: (1) the L2T framework substantially improves linguistic competence up to 11.3\% (2.8\% on average) on BLiMP~\cite{warstadt-etal-2020-blimp-benchmark}, while retaining general performance; and (2) empirical evidence that L2T accelerates this process.

\begin{table}[!t]
\scriptsize
\centering
\renewcommand{\arraystretch}{0.89}
\resizebox{0.96\columnwidth}{!}{
\begin{tabularx}{\columnwidth}{@{}c@{\hspace{5pt}}X@{\hspace{5pt}}c@{}}
\toprule
\textbf{Type} & \textbf{Tasks and Examples} &\textbf{Links} \\ \midrule
\textbf{Char} & \textbf{Char Count}: count characters (Text $\to$ 4) & \S\ref{char_count}\\
& \textbf{Masked Char}: reconstruct masked characters (c\_ar $\to$ char) & \S\ref{masked_char}\\
& \textbf{Space}: restore whitespace (Ilikea $\to$ I like a) & \S\ref{space}\\
& \textbf{Typo}: correct synthetic typos (typ0 $\to$ typo) & \S\ref{typo}\\
\midrule
\textbf{Word} & \textbf{Last}: predict concluding phrase ([Text]$_{\text{A/B}}$ $\to$ A) & \S\ref{last}\\
& \textbf{Masked Word}: reconstruct tokens (I [MASK] $\to$ I am) & \S\ref{masked_word}\\
& \textbf{Random}: correct tokens (Sea am $\to$ I am) & \S\ref{random}\\
& \textbf{Shuffle}: restore word order ($w_1 w_3 w_2$ $\to$ $w_1 w_2 w_3$) & \S\ref{shuffle}\\
& \textbf{Token Type}: count linguistic categories ([Text]$_{\text{Digit}}$ $\to$ $n \in \mathbb{Z}$) & \S\ref{token_type}\\
\midrule
\textbf{Sent} & \textbf{Deletion}: remove unrelated sentence (A [X] C $\to$ AC) & \S\ref{deletion}\\
& \textbf{Reordering}: restore sentence flow ($S_3 S_1 S_2$ $\to$ $S_1 S_2 S_3$) & \S\ref{reordering}\\
\midrule
\textbf{Disc} & \textbf{Fill Middle}: fill-in-the-middle for passages ($P_1$ ? $P_3$ $\to$ $P_2$) & \S\ref{fill_middle}\\
& \textbf{Half}: complete the latter half ([Start]... $\to$ [End]) & \S\ref{half}\\
& \textbf{One}: generate from one-word prefix ([Word] $\to$ [Text]) & \S\ref{one} \\ \bottomrule
\end{tabularx}
}
\caption{The 14 L2T tasks with links to the detailed task definitions and examples in Appendix \ref{appendix:sample}.}
\label{tab:tasks}
\end{table}

\section{L2T: Language Learning Tasks} 
\label{sec:tasks}

The L2T framework comprises 14 language learning tasks (Table \ref{tab:tasks}) designed to provide explicit structural stimulation. Each task converts a raw text segment into a structured input-output pair $(x, y)$, where $x$ is a linguistically perturbed or queried input, and $y$ is the restored or analyzed output.

While prior work has often relied on architectural modifications~\cite{xu-etal-2021-syntax} or complex curricula~\cite{hu-etal-2025-circuits,oba-etal-2023-babylm,mi-2023-mmi01}, our aim is distinct: \textit{we investigate how structured stimulation impacts pre-training dynamics and the development of linguistic competence}. We hypothesize that the \textit{rote learning} of LMs stems in part from the single-task nature of CLM, which prioritizes surface-level statistics over structural understanding. In contrast, humans do not acquire language by optimizing a single objective; they learn through multiple tasks~\cite{spelke2022babies}.

To mirror this multi-task learning, L2T generates various tasks by leveraging raw text across four levels of linguistic granularity, providing supervision signals without external resources.
Prior work \cite{feedback_morph} shows that error correction in human learners improves morphological awareness, particularly for previously encountered forms. Accordingly, we use character-level (Char) tasks (e.g., \textit{Typo}) to target subword features, discouraging surface-level matching while enhancing morphological awareness. Word- and sentence-level tasks (e.g., \textit{Shuffle}) disrupt linear order, promoting structural inference over sequential statistics, consistent with word-reordering studies \citep{AKHTAR_1999,chomsky2002syntactic,zhang_word_order}. Finally, discourse-level (Disc) tasks (e.g., \textit{Fill Middle}) require completion across longer context, supporting global coherence and ambiguity resolution. Similar completion tasks benefit human language learners \cite{keating_2008}.
By integrating these diverse signals, L2T establishes the structural scaffolding required for linguistic competence, complementing world knowledge acquired through standard CLM.

\section{Experimental Setup}

\subsection{Pre-training Data Scenarios}

We derive our pre-training corpora from the English FineWeb-Edu dataset~\cite{penedo2024the}. 
To evaluate the L2T framework under different resource constraints, we establish two distinct data scenarios (\textbf{Disjoint} and \textbf{Shared}) to determine if L2T yields consistent benefits across varying volumes of unique source text.
Following \citet{cheng-etal-2024-instruction}, we fix the total training budget at 100B tokens. 
Crucially, this budget exceeds the compute-optimal thresholds defined by the Chinchilla scaling laws~\cite{hoffmann2022trainingcomputeoptimallargelanguage}, enabling us to evaluate performance in a regime characterized by sufficient token availability relative to model size.

\begin{table*}[!t]
\centering
\tiny
\renewcommand{\arraystretch}{0.8}
\setlength{\aboverulesep}{1.3pt}
\setlength{\belowrulesep}{1.3pt}
\resizebox{\textwidth}{!}{
    \begin{tabular}{lllccccccccccccc}
    \toprule
    & & & \multicolumn{2}{c}{\textbf{Semantics}} 
    & \multicolumn{4}{c}{\textbf{Morphology}}
    & \multicolumn{6}{c}{\textbf{Syntax}}\\

    \cmidrule(lr){4-5} \cmidrule(lr){6-9} \cmidrule(lr){10-15}
    
    & \multicolumn{2}{c}{\textbf{Data}}
    & \textbf{Quant} 
    & \textbf{NPI}
    & \textbf{Ana Agr}
    & \textbf{Irregul}
    & \textbf{DN Agr}
    & \textbf{SV Agr}
    & \textbf{Arg Str}
    & \textbf{Bind} 
    & \textbf{Ctrl Rais}   
    & \textbf{Ellips} 
    & \textbf{Fill Gap}
    & \textbf{Island}
    & \textbf{Overall}\\
    \midrule
    \multirow{5}{*}{\rotatebox[origin=c]{90}{\parbox[c]{0.75cm}{\centering \tiny 500M}}} 

    & \multirow{2}{*}{\rotatebox[origin=c]{90}{\parbox[c]{0.4cm}{\centering \tiny Disj.}}}
& Raw & 65.6 & 76.5 & 93.8 & 91.8 & 93.1 & 86.0 & 78.2 & 72.8 & 78.5 & 87.1 & 77.5 & 63.0 & 78.6\\
& &L2T & \cellcolor{green!20}71.4 & \cellcolor{green!20}78.3 & \cellcolor{green!20}96.0 & \cellcolor{green!20}93.2 & 92.4 & \cellcolor{green!20}88.6 & \cellcolor{green!20}78.6 & \cellcolor{green!20}75.5 & \cellcolor{green!20}79.5 & 86.0 & 75.4 & \cellcolor{green!20}70.8 & \cellcolor{green!20}80.2\\

\cmidrule{2-16}
    & \multirow{2}{*}{\rotatebox[origin=c]{90}{\parbox[c]{0.4cm}{\centering \tiny Shar.}}}

& Raw & 67.5 & 68.8 & 94.0 & 92.3 & 93.6 & 86.5 & 78.6 & 77.9 & 79.6 & 88.5 & 75.6 & 60.6 & 78.1\\
& & L2T & \cellcolor{green!20}74.9 & \cellcolor{green!20}78.7 & \cellcolor{green!20}97.9 & \cellcolor{green!20}96.2 & 93.4 & \cellcolor{green!20}88.4 & \cellcolor{green!20}80.5 & 74.2 & \cellcolor{green!20}81.4 & 86.2 & \cellcolor{green!20}76.4 & \cellcolor{green!20}68.1 & \cellcolor{green!20}80.9\\

\midrule
\multirow{4}{*}{\rotatebox[origin=c]{90}{\parbox[c]{0.4cm}{\centering \tiny 1B}}}

& \multirow{2}{*}{\rotatebox[origin=c]{90}{\parbox[c]{0.4cm}{\centering \tiny Disj.}}}
& Raw & 71.1 & 78.6 & 95.3 & 89.5 & 94.5 & 84.8 & 78.0 & 78.6 & 79.3 & 86.7 & 75.4 & 60.2 & 79.0\\

& & L2T & \cellcolor{green!20}75.2 & 77.9 & \cellcolor{green!20}96.2 & \cellcolor{green!20}92.7 & 93.3 & \cellcolor{green!20}88.3 & \cellcolor{green!20}79.4 & 76.1 & \cellcolor{green!20}80.8 & 86.5 & 74.9 & \cellcolor{green!20}71.5 & \cellcolor{green!20}80.8\\

\cmidrule{2-16}
& \multirow{2}{*}{\rotatebox[origin=c]{90}{\parbox[c]{0.4cm}{\centering \tiny Shar.}}}
& Raw & 74.8 & 71.2 & 95.7 & 90.8 & 94.6 & 87.4 & 78.2 & 75.7 & 78.5 & 86.5 & 77.0 & 61.7 & 78.9\\
& & L2T & 72.1 & \cellcolor{green!20}82.0 & \cellcolor{green!20}97.0 & \cellcolor{green!20}94.2 & 92.7 & \cellcolor{green!20}88.4 & \cellcolor{green!20}80.0 & \cellcolor{green!20}76.7 & \cellcolor{green!20}82.0 & 84.5 & 77.0 & \cellcolor{green!20}68.6 & \cellcolor{green!20}81.2\\

    \bottomrule
    \end{tabular}
    }
    \caption{Linguistic competence on BLiMP. \colorbox{green!20}{Green} highlights better performance of L2T (ours) over Raw.
}
    \label{tab:blimp_result_merged}
\end{table*}

\paragraph{Disjoint {\rm (}Abundant Data{\rm )}.}
This configuration simulates a regime where high-quality source text is plentiful.
We partition an initial 100B token sample into two mutually exclusive subsets of equal size. The first subset is retained for standard CLM, while the second is used for generating L2T samples.
As the combination of raw and L2T samples exceeds the budget, we subsample the mixture to adhere to the 100B token limit.
The resulting Disjoint dataset comprises approximately 36B raw and 64B L2T tokens.
This setup evaluates the impact of the framework when the model has access to high document diversity alongside structured tasks.

\paragraph{Shared {\rm (}Limited Data{\rm )}.}
This configuration targets settings where source text availability is constrained.
We use a fixed set of 42B source tokens both as raw text for CLM and L2T sample generation.
This combination results in a total of 100B tokens.\footnote{While multiple transformations are possible, we assign a single, random L2T task to each sample for brevity.}
This strategy decouples the impact of task structure from data diversity, testing if L2T can improve linguistic inductive bias without the introduction of unique new documents.

\subsection{Training Configuration}

We pre-train from scratch 500M and 1B Qwen2.5-based models~\cite{qwen2025qwen25technicalreport}.
These models use the Mistral~\cite{jiang2023mistral7b} tokenizer with a 32K English-centric vocabulary. This replaces the 130K vocabulary of Qwen2.5 to reduce compute cost.
We set the sequence length to 2,048 and batch size to 256.
To ensure a fair comparison, we compute loss on all tokens.

\subsection{Evaluation Framework}

\paragraph{Baselines.}
We pre-train models (\textbf{Raw}; 500M and 1B) on 100B and 42B raw tokens for the Disjoint and Shared settings, respectively.
For fair comparison, we use a budget of 100B tokens in all settings.
Hence, the Shared Raw model trains over the same 42B tokens multiple times to meet this quota.\footnote{
Baselines and L2T models use the same hyperparameters (Appendix \ref{appendix:implementation}).}

\paragraph{Linguistic Competence.}
We use BLiMP to measure linguistic competence following \citet{shah-etal-2024-development}.
This benchmark covers 12 linguistic phenomena across 67 datasets within: Semantics, Morphology, and Syntax.
Appendix \ref{appendix:blimp_details} provides details and their associated linguistic constraints.
We prioritize zero-shot log-likelihood comparisons of minimal pairs to avoid prompting and probing biases~\cite{hu-levy-2023-prompting,alajrami-aletras-2022-pre,belinkov-2022-probing,levy-etal-2023-probing}.

\paragraph{General Benchmarks.}
We use: Reading Comprehension (\textbf{RC}; RACE, SciQ, LogiQA); Commonsense Reasoning (\textbf{CR}; ARC-Easy, COPA, OpenBookQA, SIQA, HellaSwag); and Language Modeling (LAMBADA).
We also include \textbf{ReCoRD}, which combines RC and CR.\footnote{Evaluation metrics include normalized accuracy for ARC, HellaSwag, LogiQA, OBQA, and SciQ; F1 for ReCoRD; and standard accuracy for the remaining tasks. We apply five-shot prompting for ARC, LogiQA, OBQA, and SIQA, while using zero-shot for the rest. We use a single deterministic run.\looseness=-1}
Since the volume of raw text remains constant across settings, this aims to confirm whether L2T tasks are complementary or being detrimental to CLM on raw text.

\section{Results} 
\label{sec:results}

\subsection{Linguistic Competence} \label{subsec:results-linguistic}

Table \ref{tab:blimp_result_merged} shows the performance on BLiMP for all models.
L2T consistently enhances linguistic competence.
In the Disjoint setup, L2T models outperform the Raw baselines at both 500M and 1B scales, achieving overall scores of 80.2 (500M) and 80.8 (1B), compared to Raw baseline scores of 78.6 and 79.0.
Notably, this performance gap widens in the Shared scenario, where L2T models surpass the Raw baselines by 2.8 and 2.3 points.
This reinforces our human learning analogy (\S\ref{sec:tasks}): while Shared Raw mimics ``rote learning,'' L2T mirrors multi-task acquisition by applying diverse structural lenses to the same data.
Our results confirm that linguistic competence depends not just on data volume, but on the diversity of signals applied during training.\looseness=-1

\begin{table*}[t]
    \tiny
    \centering
    \renewcommand{\arraystretch}{0.8}
    \setlength{\aboverulesep}{1.3pt}
    \setlength{\belowrulesep}{1.3pt}
    \resizebox{\linewidth}{!}{%
    \begin{tabular}{lllccccccccccc}
    \toprule

    & & & \multicolumn{3}{c}{\textbf{Reading Comprehension}} & \textbf{RC+CR} & \multicolumn{6}{c}{\textbf{Commonsense Reasoning}} & \textbf{Language Modeling}\\
    \cmidrule(lr){4-6} \cmidrule(lr){7-7} \cmidrule(lr){8-13} \cmidrule(lr){14-14}
    
    \multicolumn{3}{c}{\textbf{Data}} & 
    \textbf{RACE} & 
    \textbf{SciQ} & 
    \textbf{LogiQA} &
    \textbf{ReCoRD} &
    \textbf{ARC} & 
    \textbf{COPA} & 
    \textbf{OBQA} & 
    \textbf{PIQA} & 
    \textbf{SIQA} & 
    \textbf{HellaSwag} & 
    \textbf{LAMBADA}\\
    \midrule

    \multirow{5}{*}{\rotatebox[origin=c]{90}{\parbox[c]{0.75cm}{\centering \tiny 500M}}}
    
    & \multirow{2}{*}{\rotatebox[origin=c]{90}{\parbox[c]{0.4cm}{\centering \tiny Disj.}}}
    & Raw & 30.0 & 67.9 & 23.8 & 62.9 & 57.4 & 66.0 & 31.4 & 63.9 & 38.1 & 37.8 & 30.6\\
    & & L2T & 29.8 & 67.3 & \cellcolor{green!20}26.6 & 60.7 & 56.4 & 64.0 & 30.8 & 62.9 & \cellcolor{green!20}38.3 & 35.2 & 28.2\\

    \cmidrule{2-14}
    
    & \multirow{2}{*}{\rotatebox[origin=c]{90}{\parbox[c]{0.4cm}{\centering \tiny Shar.}}}
    & Raw & 28.3 & 66.1 & 25.0 & 62.2 & 56.6 & 66.0 & 29.0 & 64.0 & 38.1 & 37.1 & 30.6\\
    & & L2T & \cellcolor{green!20}29.5 & 63.0 & \cellcolor{green!20}26.9 & 62.0 & \cellcolor{green!20}57.7 & 64.0 & \cellcolor{green!20}29.6 & \cellcolor{green!20}65.1 & \cellcolor{green!20}38.5 & 36.2 & 30.2\\

    \midrule
    \multirow{4}{*}{\rotatebox[origin=c]{90}{\parbox[c]{0.5cm}{\centering \tiny 1B}}}
    & \multirow{2}{*}{\rotatebox[origin=c]{90}{\parbox[c]{0.4cm}{\centering \tiny Disj.}}}
    & Raw & 29.6 & 72.4 & 24.4 & 64.9 & 60.4 & 61.0 & 31.0 & 65.9 & 38.6 & 39.7 & 32.7\\
    & & L2T & \cellcolor{green!20}29.8 & 70.2 & \cellcolor{green!20}27.2 & 63.1 & 58.4 & \cellcolor{green!20}66.0 & \cellcolor{green!20}32.0 & 65.9 & 38.5 & 37.8 & 30.9\\
    
    \cmidrule{2-14}
    & \multirow{2}{*}{\rotatebox[origin=c]{90}{\parbox[c]{0.4cm}{\centering \tiny Shar.}}}
    & Raw & 29.5 & 68.0 & 26.1 & 65.3 & 60.6 & 69.0 & 31.2 & 66.3 & 38.4 & 39.4 & 31.9\\
    & & L2T & \cellcolor{green!20}30.2 & 68.0 & \cellcolor{green!20}26.7 & 63.6 & 56.4 & 65.0 & 30.0 & 64.3 & 37.4 & 37.6 & 31.3\\

    \midrule
    \rowcolor{gray!20}\multicolumn{3}{c}{\tiny Random guessing} & 25.0 & 25.0 & 25.0 & 19.1 & 25.0 & 50.0 & 25.0 & 50.0 & 33.3 & 25.0 & 0.0\\

    \bottomrule
    \end{tabular}%
    }
    \caption{General benchmark performance. \colorbox{green!20}{Green} denotes better performance of L2T (ours) over Raw.}
    \label{tab:downstream_task_result_merged}
\end{table*}

We further analyze the effects across different linguistic subfields and phenomena.
L2T consistently enhances linguistic competence across model sizes and data scenarios, improving six phenomena across all subfields.
Island effects (Island) exhibit the most substantial gains, ranging from 6.9 (1B; Shared) to 11.3 points (1B; Disjoint).
Detailed analysis (\S\ref{sec:analysis} and Appendix \ref{appendix:analysis-task}) reveals that nearly all L2T tasks contribute to this gain.
This suggests that structured tasks across varying granularity, rather than solely local transformations, provide a structural scaffolding for capturing long-distance dependencies.\looseness=-1

Conversely, L2T offers no improvement for determiner-noun agreement (DN Agr) or ellipsis (Ellips).
These results serve as a diagnostic of our method: performance for these phenomena appears to reach saturation, as Raw models achieve high accuracy (93+ and 86+, respectively) through implicit learning of frequent patterns~\cite{ellipsis,khullar-etal-2020-noel}.
This result aligns with the finding of \citet{shah-etal-2024-development}, who report that LMs are substantially accurate on morphological tasks followed by syntactic and semantic tasks.
However, limits to this scaffolding remain. 
Complex phenomena, such as Filler-Gap dependencies (Fill Gap), show no improvement from individual tasks (see Appendix \ref{appendix:analysis-task}), indicating that capturing such structures is more challenging and may require targeted discourse-level objectives.

\begin{figure}[!t]
\centering
\includegraphics[width=\columnwidth]{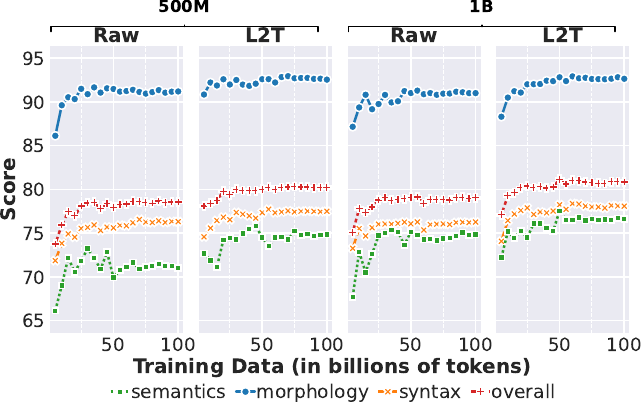}
\caption{Accuracy by linguistic subfield in BLiMP between Raw and L2T across model sizes and training steps using Disjoint Raw and L2T data.}
\label{fig:blimp_subfield}
\end{figure}

Finally, Figure \ref{fig:blimp_subfield} illustrates the development of linguistic competence throughout the pre-training process.
Across subfields, substantial improvements occur during the initial 20-30B tokens of training, with gains continuing and leveling off around 50B tokens. These results reveal similar trajectories as those identified by \citet{shah-etal-2024-development} for Pythia models trained on raw text.
However, L2T models consistently outperform Raw models across all linguistic subfields from the earliest stages of training, evident as early as 5B tokens.
This indicates that L2T effectively boosts learning in ``the window of maximal development'', the period where the model improves its cognitive abilities linearly~\cite{shah-etal-2024-development}.
At this 5B token mark, L2T models show an average performance advantage over Raw models ranging from 3.3 (Syntax) to 6.5 (Semantics) for the 500M scale, and from 1.0 (Syntax) to 4.5 (Semantics) for the 1B scale.
This faster acquisition indicates that L2T tasks function as a force multiplier for efficiency by introducing linguistic inductive biases complementary to CLM.
By effectively stimulating language learning, L2T establishes a durable advantage, preserving a performance gap that the implicit acquisition of standard Raw models cannot close.

\subsection{General Benchmarks} \label{subsec:results-general}

Table \ref{tab:downstream_task_result_merged} compares the performance of L2T and Raw models on general benchmarks.

\paragraph{Disjoint.}
In the Disjoint setup, L2T models maintain competitive performance relative to Raw.
The 500M model shows an average drop of 0.87 points, while the 1B model shows a negligible average difference of 0.07 points.
This confirms that general reasoning performance remains stable when the model retains access to a large volume of unique documents.\looseness=-1

\paragraph{Shared.}
In the Shared configuration, the impact of L2T varies by model scale.
While the 500M model achieves an average improvement of 0.15 points, the 1B model experiences a performance drop of 1.38 points, primarily in CR tasks such as ARC (-4.2).
This divergence highlights a tension between linguistic structure learning and factual reinforcement \citep{FedorenkoVarley2016LanguageThought}.
The Shared Raw baseline encounters the same 42B tokens multiple times, which likely reinforces the retention of factual world knowledge.
In contrast, the L2T setup replaces a portion of these repetitions with structured tasks.
For the 1B model, the results indicate that the benefits of linguistic stimulation do not fully compensate for the reduced exposure to raw sequences in knowledge-intensive benchmarks.
Thus, to perform well on these benchmarks, larger models need to balance structural induction with factual consolidation.\footnote{The analysis in Appendix \ref{appendix:analysis-mix} supports this conclusion by showing that performance on knowledge-intensive benchmarks is sensitive to the ratio of raw text.}

\begin{figure}[!t]
\centering
\includegraphics[width=\columnwidth]{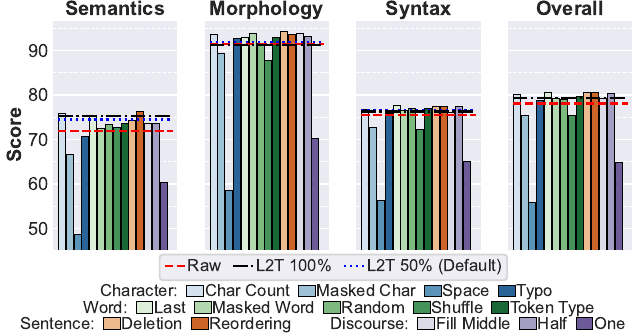}
\caption{
Linguistic competence comparisons on BLiMP between different L2T models trained on specific 25B token single task data.
}
\label{fig:single_blimp_subfield}
\end{figure}

\section{Analysis}
\label{sec:analysis}

\paragraph{Efficacy of Individual Tasks.}
To isolate the contributions of specific tasks, we pre-train models on 25B task-specific tokens.
As shown in Figure \ref{fig:single_blimp_subfield}, nine tasks (e.g., \textit{Char Count}, \textit{Reordering}) consistently outperform the Raw baseline, suggesting they provide critical structural scaffolding for linguistic acquisition.\footnote{See Appendix \ref{appendix:analysis-task} for the phenomenon-level breakdown.}
In contrast, character-level tasks such as \textit{Space} and \textit{Masked Char} underperform the Raw baseline, with a performance degradation of up to 33 points in Morphology for the Space task.
We speculate that these tasks might create an unstable training signal when used standalone.
Furthermore, while the \textit{One} task largely preserves general performance due to the functional overlap of the objective with standard CLM (Figure \ref{fig:single_downstream}), it provides insufficient structural signal to elicit linguistic gains.
Ultimately, the combined L2T framework consistently surpasses the performance of most single-task models on both BLiMP and general benchmarks.
This underscores the robustness of the approach and demonstrates that the integration of diverse tasks elicits complementary strengths, a result that parallels the generalization gains observed when fine-tuning LMs on broad task distributions \citep{wei2022finetuned, padmakumar-etal-2022-exploring, sanh2022multitask}.

\begin{table}[t]
\centering
\renewcommand{\arraystretch}{0.8}
\setlength{\aboverulesep}{1.3pt}
\setlength{\belowrulesep}{1.3pt}
\resizebox{\linewidth}{!}{
\begin{tabular}{@{}l@{\hspace{0.5em}}l@{\hspace{0.5em}}l cc cc c}
\toprule
\multicolumn{3}{c}{\textbf{Data}} & \multicolumn{2}{c}{\textbf{Numeric Abilities}} & \multicolumn{2}{c}{\textbf{Conceptual Understanding}} & \textbf{Fluid Reasoning} \\
\cmidrule(lr){4-5} \cmidrule(lr){6-7} \cmidrule(lr){8-8}
& & & \textbf{Distance} & \textbf{Ratio} & \textbf{Latent Rep.} & \textbf{Zero-Shot} & \textbf{RPM} \\
\midrule
\multirow{4}{*}{\rotatebox{90}{500M}} & \multirow{2}{*}{\rotatebox[origin=c]{90}{\centering Disj.}} & Raw & 0.958 & 0.826 & -0.080 & 0.172 & 0.312 \\
& & L2T & \textbf{0.961} & \textbf{0.830} & -0.098 & \textbf{0.184} & \textbf{0.366} \\
\cmidrule{2-8}
& \multirow{2}{*}{\rotatebox[origin=c]{90}{\centering Shar.}} & Raw & 0.960 & 0.778 & -0.074 & 0.134 & 0.292 \\
& & L2T & 0.958 & \textbf{0.847} & -0.087 & \textbf{0.201} & \textbf{0.300} \\
\midrule
\multirow{4}{*}{\rotatebox{90}{1B}} & \multirow{2}{*}{\rotatebox[origin=c]{90}{\centering Disj.}} & Raw & 0.947 & 0.797 & -0.128 & 0.164 & 0.322 \\
& & L2T & \textbf{0.969} & \textbf{0.870} & \textbf{-0.125} & \textbf{0.201} & \textbf{0.352} \\
\cmidrule{2-8}
& \multirow{2}{*}{\rotatebox[origin=c]{90}{\centering Shar.}} & Raw & 0.970 & 0.833 & -0.125 & 0.137 & 0.322 \\
& & L2T & \textbf{0.973} & \textbf{0.850} & -0.125 & \textbf{0.173} & 0.298 \\
\bottomrule
\end{tabular}
}
\caption{Psychometric evaluation results across cognitive domains. Performance is measured using $R^2$ for Numeric Abilities, Spearman $\rho$ for Conceptual Understanding, and Accuracy for Fluid Reasoning.
\textbf{Bold} indicates the superior result for a given scale and data regime.\looseness=-1}
\label{tab:psychometric_results}
\end{table}

\paragraph{Generalization beyond Linguistic Competence.}
We evaluate the impact of L2T on broader cognitive intelligence using the psychometric framework of \citet{shah-etal-2024-development}, covering \textit{Numeric Abilities}, \textit{Conceptual Understanding}, and \textit{Fluid Reasoning}.
Table \ref{tab:psychometric_results} shows that structural stimulation yields consistent gains beyond linguistic competence.
For fluid reasoning, L2T substantially enhances Raven Progressive Matrices (RPM) performance (+5.4\% for 500M and +3.0\% for 1B) in the Disjoint setting.
This suggests that L2T enhances the capacity of the model to induce abstract patterns, which is fundamental to fluid intelligence. 
Furthermore, L2T models consistently demonstrate a stronger Ratio Effect in numeric abilities (e.g., 0.870 for 1B L2T vs. 0.797 for 1B Raw in the Disjoint setting), aligning the representation of magnitude more closely with the processing of humans.
Finally, L2T improves conceptual understanding by enhancing zero-shot alignment with the typicality gradients of humans. 
These results suggest that diverse structured tasks enable the model to develop representations of semantic categorization rather than relying on simple word-level associations.

\paragraph{Discussion.}
The primary value of the L2T framework resides in the shift of pre-training from the reproduction of surface patterns to structural induction.
As demonstrated in Figure \ref{fig:blimp_subfield}, L2T models outperform Raw baselines in linguistic competence as early as 5B tokens, establishing the framework as an efficient mechanism for imparting linguistic inductive biases during the initial stages of pre-training.
Unlike instruction tuning, L2T induces structure directly from raw text without external supervision.
We expect that the field will adopt the L2T paradigm as a foundational objective to ensure that models grasp the underlying linguistic scaffolding.
Nonetheless, a direct comparison with supervised augmentation remains a critical direction.
Such an analysis would clarify the advantages of the unsupervised approach of L2T relative to methods that rely on human-annotated data.

\section{Conclusion}
We presented L2T, a pre-training framework that integrates language learning tasks alongside standard next-token prediction.
By necessitating the extraction and restructuring of information, these tasks demand processing beyond rote learning, effectively stimulating faster and improved linguistic competence development.
Future work will extend L2T to multilingual settings to investigate the learning behavior across languages.

\section*{Limitations}
\paragraph{Task Scope.}
The current design of the L2T framework focuses largely on constraints at the level of the sentence. While this effectively targets local syntactic and semantic dependencies, it does not explicitly address broader linguistic phenomena. Future work could expand to include more discourse-level and cross-sentence tasks. Such tasks would allow the model to capture long-range dependencies more effectively.

\paragraph{Model Scale.}
Due to constraints on computational resources, we restrict our evaluation to models at the 500M and 1B parameter scales. We follow the protocol of \citet{cheng-etal-2024-instruction} by training our models from scratch with a reasonable academic budget of 100B tokens. We acknowledge that the effects of different pre-training objectives might vary at larger scales (e.g., 10B+ parameters), and their investigation is a great opportunity that should be explored by community members that have access to the required compute.
We speculate that the structural scaffolding of L2T could be particularly beneficial during the initial stages of pre-training for larger models, as observed in \citet{shah-etal-2024-development} and \citet{hu-etal-2025-circuits}.
However, the drop in reasoning performance of the Shared 1B model suggests that larger models are likely more sensitive to the size of raw text. Consequently, the application of L2T at scale likely necessitates a careful balance between the volume of raw text and structured tasks to preserve world knowledge, and it may benefit from the use of curriculum learning (e.g., confining L2T data to the initial stage of pre-training)~\cite{10.1145/1553374.1553380,platanios-etal-2019-competence}.\looseness=-1

\paragraph{Single Seed.}
The findings in this study are based on a single training run for each configuration.
The high computational cost of pre-training models of 500M and 1B parameters from scratch for 100B tokens made multiple runs prohibitive within the constraints of the available academic infrastructure.
However, the consistency of the performance gains across two distinct model scales and two different data regimes (Shared and Disjoint) provides evidence for the robustness of the L2T framework.
This indicates that the improvements in linguistic competence result from the systematic impact of structured stimulation rather than sampling variance.
Furthermore, the hyperparameters follow the established protocols of \citet{cheng-etal-2024-instruction} to ensure standard training dynamics.
To support verification and the scaling of L2T by the research community, the source code and the data transformation pipeline are publicly available at \url{https://github.com/gucci-j/l2t}.

\paragraph{Analysis of Individual Tasks.}
Assessing the efficacy of individual tasks (\S\ref{sec:analysis} and Appendix \ref{appendix:analysis-task}) relies on experiments conducted at a reduced scale of 25B tokens.
Although this volume of data may not provide a comprehensive profile across the entire pre-training duration, it specifically covers the window of maximal development.
During this period, LMs typically demonstrate a linear improvement in cognitive and linguistic abilities~\citep{shah-etal-2024-development}.
Because performance gains tend to plateau following this initial phase, this interval represents a theoretically grounded window for the assessment of task impact.
This targeted analysis enables the identification of essential structural signals while maintaining computational feasibility.

\section*{Acknowledgements}
We would like to thank Anthony Hughes for the valuable feedback.
We acknowledge (1) the use of the University of Oxford Advanced Research Computing (ARC) facility: \url{http://dx.doi.org/10.5281/zenodo.22558} and (2) the Isambard-AI National AI Research Resource (AIRR)~\cite{mcintoshsmith2024isambardaileadershipclasssupercomputer}, which is operated by the University of Bristol and is funded by the UK Government’s Department for Science, Innovation and Technology (DSIT) via UK Research and Innovation; and the Science and Technology Facilities Council [ST/AIRR/I-A-I/1023]. AY is supported by the Engineering and Physical Sciences Research Council (EPSRC)  [grant number EP/W524360/1] and the Japan Student Services Organization (JASSO) Student Exchange Support Program (Graduate Scholarship for Degree Seeking Students). MM is supported by the UKRI AI Centre for Doctoral Training in Speech and Language Technologies (SLT) and their Applications funded by UK Research and Innovation [grant number EP/S023062/1]. NA is partly supported by the EPSRC [grant number EP/Y009800/1].

\bibliography{anthology-1,anthology-2,custom}
\bibliographystyle{acl_natbib}

\newpage
\appendix

\begin{tcolorbox}[
    title=Appendix Directory,
    colback=lightgray!10,
    colframe=black,
    fonttitle=\bfseries, 
    rounded corners
]
\begin{itemize}
    \item \textbf{Appendix A:} \hyperref[appendix:related_work]{Related Work}
    
    \item \textbf{Appendix B:} \hyperref[appendix:sample]{Details on L2T: Language Learning Tasks}
    \begin{itemize}
        \item \hyperref[appendix:char_level]{Character-level}
        \item \hyperref[appendix:word_level]{Word-level}
        \item \hyperref[appendix:sentence_level]{Sentence-level}
        \item \hyperref[appendix:discourse_level]{Discourse-level}
    \end{itemize}

    \item \textbf{Appendix C:} \hyperref[appendix:setup]{Extended Experimental Setup}
    \begin{itemize}
        \item \hyperref[appendix:data_construction]{Pre-training Data Construction}

        \item \hyperref[appendix:implementation]{Implementation and Training Details}

        \item \hyperref[appendix:blimp_details]{Evaluation Details: BLiMP Benchmark}
    \end{itemize}

    \item \textbf{Appendix D:} \hyperref[appendix:analysis-superglue]{Analysis on Additional General Benchmarks}

    \item \textbf{Appendix E:} \hyperref[appendix:analysis-task]{Efficacy of Individual Tasks}

    \item \textbf{Appendix F:} \hyperref[appendix:analysis-mix]{Mixing Ratio of Raw and L2T Data}

    \item \textbf{Appendix G:} \hyperref[appendix:qualitative]{Qualitative Analysis}
    
    \item \textbf{Appendix H:} \hyperref[appendix:license]{License}

    \item \textbf{Appendix I:} \hyperref[appendix:genai]{Use of Generative AI Tools}
\end{itemize}
\end{tcolorbox}

\section{Related Work} \label{appendix:related_work}

\subsection{Pre-training and Linguistic Competence}

Previous work has questioned the depth of linguistic understanding in LMs~\cite{rogers-etal-2020-primer,10.1145/3442188.3445922,chang-bergen-2024-language,lopezotal2025linguisticinterpretabilitytransformerbasedlanguage}.
Empirical analysis reveals that despite proficiency in generating coherent text, models often fail to process distant and complex co-occurrences, such as rhetorical relations~\cite{waldis-etal-2024-holmes}, and fine-grained linguistic annotation tasks, including noun structures at the phrase level~\cite{cheng-amiri-2025-linguistic}.
We argue that these shortcomings occur because training signals from standard CLM lack the structural scaffolding necessary for the model to move beyond surface-level statistics.

Consequently, understanding how LMs acquire linguistic knowledge during pre-training remains a key research area.
Evidence indicates that the characteristics of pre-training data play a crucial role.
For instance, child-directed language aids grammar induction more effectively than conventional text~\cite{huebner-etal-2021-babyberta,yadavalli-etal-2023-slabert}.
Likewise, pre-training on artificial language data designed to model specific structures, such as nesting dependencies, transfers knowledge successfully to natural language tasks~\cite{Chiang_Lee_2022,ri-tsuruoka-2022-pretraining,hu-etal-2025-circuits}.
Further, \citet{alajrami-aletras-2022-pre} suggest that the architecture of the model and pre-training data influence linguistic acquisition more than the objective function.
These findings collectively support the data-centric approach of this work to enhance linguistic competence by providing the explicit guidance required to resolve complex linguistic dependencies.

\subsection{Enhancing Linguistic Competence of LMs}

Prior research explores various strategies to improve the linguistic competence of LMs, typically involving architectural modifications~\cite{xu-etal-2021-syntax}, auxiliary tasks or objectives~\cite{kuncoro-etal-2019-scalable,xu-etal-2021-syntax,zhang-etal-2022-syntax,mueller-etal-2022-label,cui2022lertlinguisticallymotivatedpretrainedlanguage,guo-etal-2024-mitigating}, curriculum learning~\cite{hu-etal-2025-circuits}, or data transformation~\cite{guo-etal-2024-mitigating}.

However, research specifically targeting decoder-based LMs remains limited~\cite{lopezotal2025linguisticinterpretabilitytransformerbasedlanguage}.
Existing methods often focus on isolated phenomena or rely on external resources.
For instance, \citet{guo-etal-2024-mitigating} introduce semantic-aware permutation to mitigate the ``reversal curse,'' but this requires an auxiliary LM and focuses on continual pre-training.
Similarly, \citet{hu-etal-2025-circuits} utilize formal language data to capture hierarchical dependencies; while this improves generalization, it functions primarily as a warm-up phase using synthetic structures.

In contrast, L2T targets broad linguistic abilities in decoder-based LMs without auxiliary models or external knowledge.
By training from scratch, we isolate the impact of our data-centric intervention, demonstrating how structured tasks alone can stimulate the development of linguistic competence during pre-training.

\subsection{Self-Supervised Objectives and Data Transformation for Pre-training}
Early work explores various self-supervised objectives, particularly for encoder-based models, to improve downstream performance and computational efficiency, and to interpret learned representations~\cite{aroca-ouellette-rudzicz-2020-losses,yamaguchi-etal-2021-frustratingly,di-liello-etal-2022-effective,yamaguchi-etal-2023-task,alajrami-aletras-2022-pre,el-mesbahi-etal-2023-utility,alajrami-etal-2023-understanding}.\looseness=-1

More recently, researchers have focused on the transformation of raw text into structured input-output pairs suitable for CLM.
These methods include generation of pairs based on predefined self-supervised tasks~\cite{chen-etal-2022-improving}, curation of instruction-response pairs using auxiliary models~\cite{cheng-etal-2024-instruction}, creation of pseudo-labeled data~\cite{gu-etal-2022-learning}, or adaptation of domain text into task formats~\cite{cheng2024adapting}.
These approaches typically aim to enhance general capabilities, improve few-shot learning, or adapt models to specific tasks or domains, often through continual pre-training~\cite{chen-etal-2022-improving}.

Our work aligns with the use of data transformation and self-supervision but differs fundamentally in approach and objective.
Unlike previous strategies that rely on external models or task-specific datasets for transformation~\cite{cheng-etal-2024-instruction, gu-etal-2022-learning, cheng2024adapting}, our data transformation applies intrinsically to any raw text using predefined rules.
Crucially, while prior work often targets improved downstream task performance, our goal is distinct: \textit{to enhance the linguistic competence of LMs during pre-training.}

\section{Details on L2T: Language Learning Tasks}
\label{appendix:sample}

\begin{figure*}[!t]
\centering
\includegraphics[
    width=0.9\textwidth
]{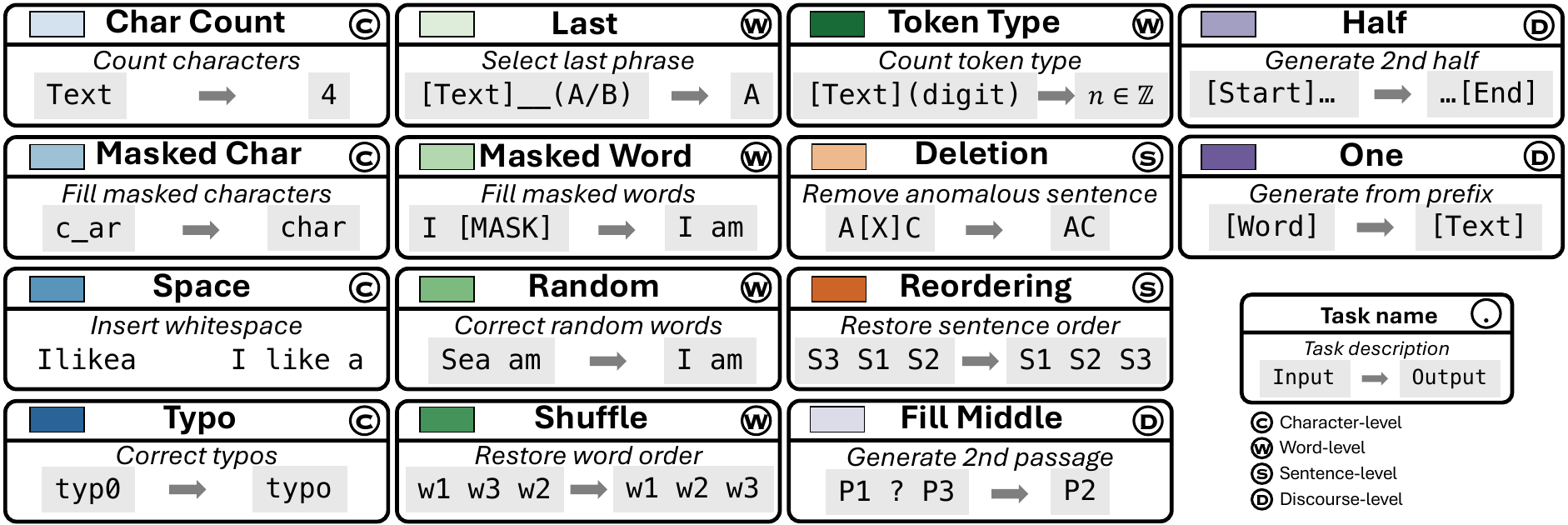}
\caption{Overview of the 14 language learning tasks.
Colors denote linguistic granularity: character (blue), word (green), sentence (orange), and discourse (purple).\looseness=-1}
\label{fig:task}
\end{figure*}

We introduce 14 language learning tasks (Figure \ref{fig:task}) designed to convert raw text into structured input-output pairs.
These tasks highlight specific linguistic aspects and necessitate processing capabilities beyond simple next-token prediction.
We organize the tasks by linguistic granularity, ranging from local surface dependencies (character/word) to deep syntactic and global coherence structures (sentence/discourse).
In the following, we include excerpts from the training samples for each task. [...] signifies that text is omitted for brevity.

\subsection{Character-level} \label{appendix:char_level}
\subsubsection{Character Count {\rm(}Char Count{\rm)}} \label{char_count}
Char Count counts total characters in the input text. This task promotes fine-grained processing and arithmetic abilities, aiding the handling of morphological variations.

\begin{examplebox}{Char Count}
Calculate the character count of the following text:\\

The younger generation can never hope to see again the forest primeval, but the next thing to seeing it with one's own eyes is to see it in Cooper's word pictures, as it was before the ax of the lumberman laid it in ruins. Of Cooper's sea stories, the best is The Pilot, which tells in graphic style of the exploits of John Paul Jones in English waters. [...] To have created Leather-stocking is a passport to enduring fame; yet Cooper added to this typical American backwoodsman Long Tom Coffin, the shrewd Yankee sailor, and a long line of other original characters.\\

(A) 1833
\end{examplebox}

\subsubsection{Masked Character Replacement {\rm(}Masked Char{\rm)}} \label{masked_char}
Masked Char reconstructs the original text after replacing a percentage of characters (e.g., 15\%) with a mask token. This fosters character-level understanding and morphological processing.

\begin{examplebox}{Masked Char}
[...] A single g+++ld aiguill++++++te is w+++rn o+++ offic+++r cut unifo+++ms of minor +++etach+++ent+++. +++ doubl+++ gol+++ aig+++il+++e+++te is wo++++++ on officer cut un+++forms of major detachment+++. All +++ersonnel of +++ajor detac+++ments wear +++f+++icer cut uni+++orms.\\

A: [...] A single silver aiguillette is worn on enlisted cut uniforms of minor detachments. A single gold aiguillette is worn on officer cut uniforms of minor detachments. A double gold aiguillette is worn on officer cut uniforms of major detachments. All personnel of major detachments wear officer cut uniforms.
\end{examplebox}

\subsubsection{Whitespace Insertion {\rm(}Space{\rm)}} \label{space}
Space restores original spacing to text stripped of whitespace. This enforces word boundary learning and tokenization rules while implicitly modeling grammatical structures defined by spacing.

\begin{examplebox}{Space}
Femalesmakeof55.5\%oftherecordeddeathsand
malesmakeup44.5\%Whatwastheaveragebirthwe
ightofthenewbornsthatsurvived?Therewerebirth
weightsrecordedfornineofthenewbornsthatsurvi
ved.[...]-Theleadingcauseofnewbornmortalityw
aspredationbypumas.Thiswasfollowedbystarvat
ion.\\

Response: Females make of 55.5\% of the recorded deaths and males make up 44.5\% What was the average birth weight of the newborns that survived? [...] - The leading cause of newborn mortality was predation by pumas. This was followed by starvation.
\end{examplebox}

\subsubsection{Typo Correction {\rm(}Typo{\rm)}} \label{typo}
Typo corrects synthetically introduced typos generated by replacing a random percentage of characters (e.g., 3-8\%). This task targets character and word recognition.

\begin{examplebox}{Typo}
For example, indiaiduals with a first-degree relative (parent, sibling or child) who was diagnosed with lung cancer at an eally age may be at increased risk. [...] "Accessiyility to qhis team df plysicians with expertiso in every aspect of diagnosis zo tdeatmlnt wyfl allow fcr more coordinated care, minumibing any scweenfng harms."\\

For example, individuals with a first-degree relative (parent, sibling or child) who was diagnosed with lung cancer at an early age may be at increased risk. [...] "Accessibility to this team of physicians with expertise in every aspect of diagnosis to treatment will allow for more coordinated care, minimizing any screening harms."
\end{examplebox}

\subsection{Word-level} \label{appendix:word_level}
\subsubsection{Last Phrase Prediction {\rm(}Last{\rm)}} \label{last}
Last selects the correct concluding phrase (the segment following the final stop word) from two candidates. Inspired by \citet{chen-etal-2022-improving}, this focuses on the understanding of sentence structure and context.

\begin{examplebox}{Last}
[...] These savings per person are converted to savings per unit area as follows: On the basis of population estimates (21) from 53 counties in New York State, the median population density was estimated at 103 persons per sq. mi (25th percentile = 67; 75th = 204). Thus, for the areas baited, the savings were calculated at \$156.56 per sq. mi for the first 2 epizootic years (\$1.52 per person x 103 persons per sq. mi), and \$30.90 per sq. mi for the post-epizootic years (\$0.30 per person x 103 persons per sq. mi). Cost-savings data from New Jersey (5) are used in +++?\\

Options:
and large estates.  
the sensitivity analysis.\\

A. the sensitivity analysis.
\end{examplebox}

\subsubsection{Masked Word Replacement {\rm(}Masked Word{\rm)}} \label{masked_word}
Masked Word reconstructs text where a percentage of words (e.g., 15\%) are replaced with a mask token. Similar to Masked Language Modeling~\cite{devlin-etal-2019-bert}, this enhances vocabulary knowledge and contextual inference.

\begin{examplebox}{Masked Word}
[...] This implies \$\$\$ +++ have a high demand for protein and require optimum temperatures for their metabolisms (()) function at optimal levels necessary for growth … Energy and protein are \_\_\_ by the organism for maintenance, growth and/or reproduction (Staton (()) al. 1986), where energy <<>> derived from carbohydrates, @@@ and fats. To prevent protein \_\_\_ +++ used as an energy source, @@@ energy should be supplied in the diet in [MASK] form of carbohydrates and fat.\\

A. [...] This implies that they have a high demand for protein and require optimum temperatures for their metabolisms to function at optimal levels necessary for growth … Energy and protein are required by the organism for maintenance, growth and/or reproduction (Staton et al. 1986), where energy is derived from carbohydrates, protein and fats. To prevent protein from being used as an energy source, sufficient energy should be supplied in the diet in the form of carbohydrates and fat.
\end{examplebox}

\subsubsection{Random Word Replacement {\rm(}Random{\rm)}} \label{random}
Random corrects text where 5\% to 10\% of words are replaced by random vocabulary tokens. Inspired by \citet{yamaguchi-etal-2021-frustratingly}, this improves robustness to noise and deepens contextual understanding.

\begin{examplebox}{Random}
We mortgage find the meaning. [...] If you read locus one theoretical book on the topic of narrative, this one is a testament candidate. While theoretical in perspective, it’s written pessimist a manner that will appeal to anyone involved in telling stories.\\

We cannot find the meaning. [...] If you read only one theoretical book on the topic of narrative, this one is a good candidate. While theoretical in perspective, it’s written in a manner that will appeal to anyone involved in telling stories.
\end{examplebox}

\subsubsection{Shuffled Word Correction {\rm(}Shuffle{\rm)}} \label{shuffle}
Shuffle restores the original order of text with 5\% to 10\% shuffled words. Also inspired by \citet{yamaguchi-etal-2021-frustratingly}, this focuses on syntax understanding and local dependencies.

\begin{examplebox}{Shuffle}
white loops and buckles permitted the The box to Side worn from either percussion waistbelt or a cartridge box strap. Marines wore a white buff of cartridge box strap that appears to men about 2 1/4 inches wide. [...] Equipped with a Maynard tape primer system, the priming system of the M1855 was similar to the cap guns used by been cross Musket slings were of Black Leather, and officers arms cautioned against "putting muskets in the hands of the be without slings." Officer be Arms There does not seem to be any standard side arm issued to Marine Officers of the Civil War era.\\

These loops and buckles permitted the cartridge box to be worn from either the waistbelt or a cartridge box strap. Marines wore a white buff leather cartridge box strap that appears to be about 2 1/4 inches wide. [...] Equipped with a Maynard tape primer system, the priming system of the M1855 was similar to the cap guns used by children today. Musket slings were of Black Leather, and officers were cautioned against "putting muskets in the hands of the men without slings." Officer Side Arms There does not seem to be any standard side arm issued to Marine Officers of the Civil War era.
\end{examplebox}

\subsubsection{Token Class Count {\rm(}Token Type{\rm)}} \label{token_type}
Token Type counts occurrences of specific word types (e.g., digits, stopwords, punctuation, content words). This encourages classification based on linguistic categories, enhancing the understanding of text composition.

\begin{examplebox}{Token Type}
Count the punctuation marks in the following passage.\\

Whereas watching a relaxation video reduces stress, lowers the levels of stress hormones in the blood stream and induces relaxation. [...] Click here to be taken to Comparison Between the Fight/Flight Response and Relaxation Response\\

A. 36
\end{examplebox}

\subsection{Sentence-level} \label{appendix:sentence_level}
\subsubsection{Sentence Deletion {\rm(}Deletion{\rm)}} \label{deletion}
Deletion identifies an anomalous sentence randomly inserted from external context. With equal probability, the model either (a) reconstructs the passage with the anomaly removed, or (b) generates only the anomalous sentence. This fosters the ability to distinguish irrelevant information and maintain discourse coherence.

\begin{examplebox}{Deletion}
For men, the formal hakama is usually made of heavy black or gray, striped silk. Those designed for women usually come in dark, solid colors, with the exception of the bright red hakama worn by women who tend the Shinto shrines in Japan. [...] Hakama are the traditional garment for priests who reside at Shinto shrines throughout Japan. They are also worn for the practice of archery and for martial arts such as kendo, karate, and aikido. (PDF) STOP, THINK, SPOT FAKE NEWS . Women wear them only for specific occasions and activities, such as graduation ceremonies and traditional Japanese sports. [...] A rustic version of the hakama was also worn by farmers and woodsmen.\\

A) (PDF) STOP, THINK, SPOT FAKE NEWS .
\end{examplebox}

\subsubsection{Sentence Reordering {\rm(}Reordering{\rm)}} \label{reordering}
Reordering restores the original order of a text containing randomly permuted sentences. This encourages understanding of discourse structure, logical flow, and coherence at the sentence level.

\begin{examplebox}{Reordering}
Assignments (see full descriptions below): 1. All too often, the closer a design is to completion, the more often and drastic the demands for change become. Demands for change may come from the audience, the client, the market, or their own artistic process. Thus, a successful designer learns to rely on an ability to improvise, to respond playfully and creatively, to work with increasingly narrow limitations and assets, while never losing sight of the overall experience goals. [...] In the first semester, the exercises focus on low- and non-technical forms of interactivity, experimenting with games, improvisational theater, fine arts, toys, interactive narrative, and real and virtual play spaces.\\

A: CTIN 534: Experiments in Interactivity I Instructors: Bernie DeKoven and Tracy Fullerton Course Description: This course is the first in a two-semester sequence which will introduce students to a variety of concepts in interactivity and teach creative techniques for conceptualizing and producing original interactive projects. [...] After the first two classes, which establish the twin focal points of playfulness and improvisation, students will: · experience the fundamentals of improvisation as it applies to both theatrical and interactive development through participation in a variety of exercises · explore fundamental principles of the psychology of fun, the sociology of playfulness and the art of improvisation · address issues of agency, empathy, control, creativity, collaboration and the natural tension that occurs between artist and audience when the audience has a part in making the art Thursday 6:30 – 9:30PM Media Lab, on the second floor of the Evaluation of student performance: a. Assignments (see full descriptions below): 1. In-class design exercises 2.
\end{examplebox}

\subsection{Discourse-level} \label{appendix:discourse_level}

\subsubsection{Second Passage Completion {\rm(}Fill Middle{\rm)}} \label{fill_middle}
Fill Middle generates the missing second passage given the first and third passages of a text. Analogous to fill-in-the-middle~\cite{bavarian2022efficienttraininglanguagemodels}, this strengthens the understanding of multi-passage structure and coherence.\looseness=-1

\begin{examplebox}{Fill Middle}
[...] The first confirmed case in Latin America was identified in Brazil on 26 February, although researchers have said there are indications that there were cases there as early as January. Coronavirus has since spread
in the world, and has had more than 10,000 confirmed deaths. Chile is reporting thousands of new coronavirus cases each day, with more than 6,000 people dead. [...]\\

(A) to every country in the region. More than 2.5 million cases have been recorded, and more than 100,000 people have died, according to the European Centre for Disease Control and Prevention. That is fewer deaths than in the US and Europe, but both cases and deaths may be under-reported. Latin America’s two most populous nations, Brazil and Mexico, have seen the highest number of deaths, more than 64,000 and 30,000 respectively. Peru has the sixth highest number of reported cases
\end{examplebox}

\subsubsection{Second Half Completion {\rm(}Half{\rm)}} \label{half}
Half generates the latter half of a given text. This encourages the learning of discourse coherence, prediction of long-range dependencies, and maintenance of narrative flow.\looseness=-1

\begin{examplebox}{Half}
[...] Your child must fast six hours before an ultrasound to provide a clear view of the organs. During this painless procedure, your child lies quietly while the radiologist moves the ultrasound probe across the abdomen. A cool jelly coats the probe to improve the ultrasound picture quality. Upper Gastrointestinal\\

A) Endoscopy Exam Compassionate nurses who are experienced in helping children through these tests will start an intravenous line and help administer medications to relax your child and relieve potential pain. During this test, a long, thin telescope -- about as wide as a pen -- is inserted into the mouth, down the esophagus, and into the stomach and intestine. Through this telescope and its television camera, the gastroenterologist can see the inner lining of the esophagus, stomach and intestine. Small biopsies can be obtained from this lining. Sensors attached to your child also monitor heart rate, blood pressure and blood oxygen levels to ensure safe conditions throughout the procedure. A typical upper gastrointestinal endoscopy takes about 20 minutes. With this test, the physician examines the inside lining of the entire length of the colon or large intestine.
\end{examplebox}

\subsubsection{One Word Prefix Generation {\rm(}One{\rm)}} \label{one}
One generates the subsequent text given a single word prefix followed by newlines. This specifically trains the initiation and structuring of generation from minimal context.

\begin{examplebox}{One}
These\\

Answer: These routes allow visitors to locate works in the Museum, prepare a visit beforehand, further their knowledge of the collection with a thematic route or discover an enjoyable way of introducing the Museum to children through themes such as animals or princesses. Access to the content is through the index of collections or index of artists. There is also the option to save and select works or articles that the user considers most important or relevant in the Favourites section, and to share content on the social networks (Facebook and Twitter) through direct links in the entries on the works. [...]
\end{examplebox}

\newpage
\section{Extended Experimental Setup} \label{appendix:setup}

This section details the construction of the pre-training data, the parameters of the implementation, and the framework used for evaluation.

\subsection{Pre-training Data Construction} \label{appendix:data_construction}
We construct our pre-training data by combining standard raw text with data generated via our L2T framework.\looseness=-1

\paragraph{Data Sampling Strategies.}
We investigate two configurations for mixing raw text and L2T data.
In the Disjoint configuration, we split source documents into two distinct, non-overlapping sets of equal size. One set is used exclusively for standard CLM and the other is transformed into L2T samples.
In the Shared configuration, we utilize the exact same source documents for both tasks. This means the pipeline processes every document twice: once as raw text and once to \textbf{stimulate} linguistic learning through L2T transformation.

\paragraph{Sample Generation Pipeline.} For standard CLM, documents are tokenized and packed continuously. For L2T, source documents undergo the following pipeline: 
\begin{enumerate} 
    \item \textbf{Segmentation:} Documents are segmented into sentences and grouped into chunks of approximately 512 tokens to ensure samples consist of complete sentences. 
    \item \textbf{Transformation:} One of the 14 tasks (\S\ref{sec:tasks}) is applied to each chunk. Pairs are formatted as \texttt{[Input]\textbackslash n\textbackslash n[Prefix] [Output]} using randomized prefixes for stylistic variation. 
    \item \textbf{Packing and Mixing:} Transformed chunks are concatenated to fill the maximum sequence length and then shuffled with raw text samples. This strategy provides the structural scaffolding necessary to optimize for linguistic competence while retaining world knowledge~\cite{cheng-etal-2024-instruction}.
\end{enumerate}

\subsection{Implementation and Training Details} \label{appendix:implementation}

\paragraph{Pre-training Data Construction.}
The hyperparameters for the curation of the L2T data are listed in Table \ref{tab:hyperparams_dataset}.
We use \href{https://github.com/microsoft/BlingFire}{Bling Fire} (v0.1.8) for efficient sentence segmentation. Following \citet{chen-etal-2022-improving}, we use varied mask tokens for the Masked Word and Masked Char tasks. The Last task utilizes the stop word list of NLTK~\cite{bird-loper-2004-nltk} to identify the final segment of the text.

\begin{table}[t]
\begin{center}
\small
\begin{tabularx}{\columnwidth}{XX}
\toprule
\textbf{Hyperparameter} & \textbf{Value} or \textbf{Description}\\
\midrule
Chunk length & 512 tokens\\
Sentence segmentation & \href{https://github.com/microsoft/BlingFire}{Bling Fire} (v0.1.8)\\
Prefix between input and output & \{\texttt{``Answer:"}, \texttt{``Response:"}, \texttt{``A:"}, \texttt{``(A)"}, \texttt{``A)"}, \texttt{``A."}, \texttt{``"}\}\\
Mask token variations & \{\texttt{"[MASK]"}, \texttt{"\_\_\_"}, \texttt{"@@@"}, \texttt{"\#\#\#"}, \texttt{"+++"}, \texttt{"<<>>"}, \texttt{"(())"}, \texttt{"\$\$\$"}\}\\
Masking ratio for Masked Word & 0.15\\
Masking ratio for Masked Char & 0.15\\
Replacement ratio for Random & \texttt{uniform(0.05, 0.1)}\\
Shuffling ratio for Shuffle & \texttt{uniform(0.05, 0.1)}\\
Typo ratio for Typo & \texttt{uniform(0.01, 0.08)}\\
False sample for Last \& Deletion & Random sampling from previous document\\
\bottomrule
\end{tabularx}%
\caption{Hyperparameters for curating our TL2T pre-training data.}
\label{tab:hyperparams_dataset}
\end{center}
\end{table}

For the Token Type task, each word is classified as ``stopword'', ``digit'', or ``content'' via a prioritized procedure: (i) cleaning punctuation and symbols; (ii) matching the lowercase form against the stopword list; and (iii) verifying if the remaining string consists entirely of digits. Any tokens not meeting these criteria are classified as ``content''. To count punctuation marks, we utilize the regex pattern: \verb|!"#$%&'()*+,-./:;<=>?@[\]^_`{||\verb|}~|.

To ensure the disambiguation of tasks, specifically to distinguish Token Type from Char Count, a task-specific instruction is inserted randomly at either the beginning or the end of the input, separated by two newline characters (\texttt{\textbackslash n\textbackslash n}).

\paragraph{Pre-training Details.}
Table \ref{tab:hyperparams_model} lists the hyperparameters and configuration settings used for pre-training.

\begin{table*}[t]
\begin{center}
\small
\begin{tabularx}{\textwidth}{lXX}
\toprule
\textbf{Hyperparameters} & \textbf{500M} & \textbf{1B}\\
\midrule
Hidden size & 1024 & 1728\\
Intermediate size & 4864 & 4752\\
Max window layers & 24 & 30\\
Number of attention heads & 24 & 30\\
Number of hidden layers & 24 & 32\\
Number of key value heads & 2 & 4\\
Rope theta & 1,000,000 & 1,000,000\\
RMS norm eps & 1e-06 & 1e-06\\
Attention dropout & 0.0 & 0.0\\
Tie word embeddings & True & True\\
Hidden activation & SiLU & SiLU\\
Initializer range & 0.02 & 0.02\\
Vocabulary size & 32,000 & 32,000\\
Tokenizer & Mistral & Mistral\\
Batch size & 256 & 256\\
Train steps  & 200K (for 100B token training), 50K (for 25B token training) & 200K\\
Sequence length & 2,048 & 2,048\\
Maximum Learning Rate & 3e-4 & 3e-4\\
Learning rate scheduler & cosine & cosine\\
Warmup steps & 2,000 (for 100B token training), 1,000 (for 25B token training) & 2,000\\
Adam $\epsilon$ & 1e-8 & 1e-8\\
Adam $\beta_1$ & 0.9 & 0.9\\
Adam $\beta_2$ & 0.999 & 0.999\\
Gradient clipping & 1.0 & 1.0\\
Weight decay & 0.1 & 0.1\\
Training precision & BF16 & BF16\\
Computing infrastructure & 2 AMD MI300X GPUs (for 100B token training), 2 H100 (96GB) HBM3 GPUs (for 25B token training) & 4 H100 (96GB) HBM3 GPUs\\
Run time for each model & 16 days (for 100B token training), 3.5 days (for 25B token training) & 12 days\\
\bottomrule
\end{tabularx}%
\caption{Hyperparameters and training costs for each model scale.}
\label{tab:hyperparams_model}
\end{center}
\end{table*}

\paragraph{Libraries.}
We preprocess datasets with Hugging Face Datasets~\cite[v3.2.0]{lhoest-etal-2021-datasets}.
We use PyTorch~\cite[v2.3.0]{10.1145/3620665.3640366}, FlashAttention-2~\cite[v2.7.3]{dao2023flashattention2fasterattentionbetter}, and Hugging Face Transformers~\cite[v4.49.0]{wolf-etal-2020-transformers} for pre-training.
For evaluation, we use \texttt{lm-evaluation-harness}~\cite[v0.4.8]{eval-harness}.

\subsection{Evaluation Details: BLiMP Benchmark} \label{appendix:blimp_details}
To measure the linguistic competence of models, we utilize the BLiMP benchmark, which comprises 67 datasets covering 12 linguistic phenomena. Each sample consists of pairs of minimally different sentences that contrast in grammatical acceptability to isolate specific phenomena in semantics, morphology, or syntax.

\textbf{Semantics} includes two phenomena: quantifiers (Quant), which test restrictions on distribution (e.g., \textit{fewer than} versus \textit{at most}), and negative polarity items licensing (NPI). The latter assesses whether the model possesses knowledge regarding the accurate placement of words such as \textit{any} (e.g., \textit{I did not see anyone.}).

\textbf{Morphology} covers four phenomena: anaphora agreement (Ana Agr), which verifies whether pronouns correctly match the intended referent (e.g., \textit{John saw himself.}); irregular forms (Irregul), which evaluate knowledge of unpredictable word forms (e.g., \textit{go} becomes \textit{went}); determiner-noun agreement (DN Agr), which tests the numerical correspondence between words such as \textit{this} or \textit{these} and nouns; and subject-verb agreement (SV Agr), which confirms that the verb form agrees with the subject (e.g., \textit{he runs} versus \textit{they run}).

Finally, \textbf{Syntax} covers six phenomena: argument structure (Arg Str), which examines the combination of verbs with necessary components (e.g. \textit{eat} must accompany something to be eaten); binding (Bind), , which tests the structural relationship between a pronoun and the antecedent (e.g. \textit{John saw himself} vs. \textit{John saw him}); control/raising (Ctrl Rais), which evaluates syntactic and semantic differences between various predicate types; ellipsis (Ellips), which measures whether expressions can be omitted from a sentence; filler-gap (Fill Gap), which assesses dependencies arising from phrasal movement; and island effects (Island), which check constraints on moving sentence elements out of certain grammatical constructions.

\begin{figure*}[bth]
\centering
\includegraphics[width=\textwidth]{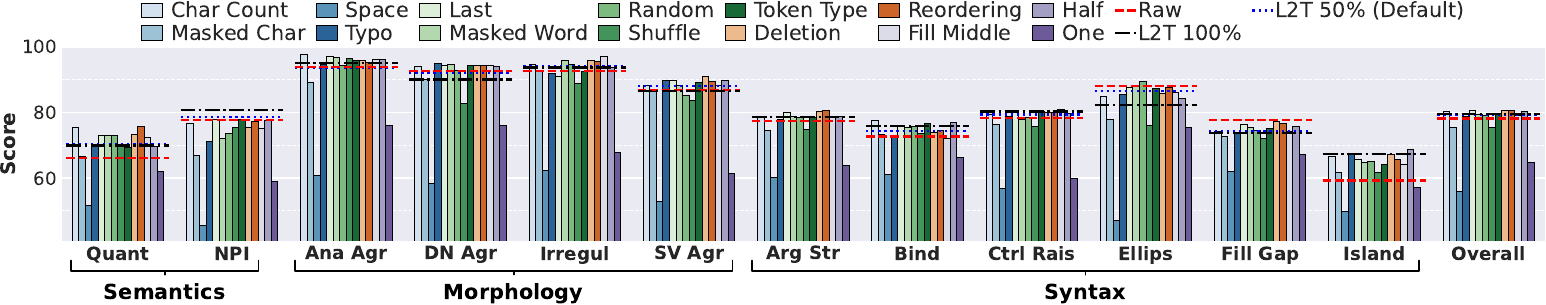}
\caption{
Linguistic competence comparisons on BLiMP between different L2T models trained on specific 25B token single task data.
}
\label{fig:single_blimp}
\end{figure*}

\begin{figure*}[bth]
\centering
\includegraphics[width=\textwidth]{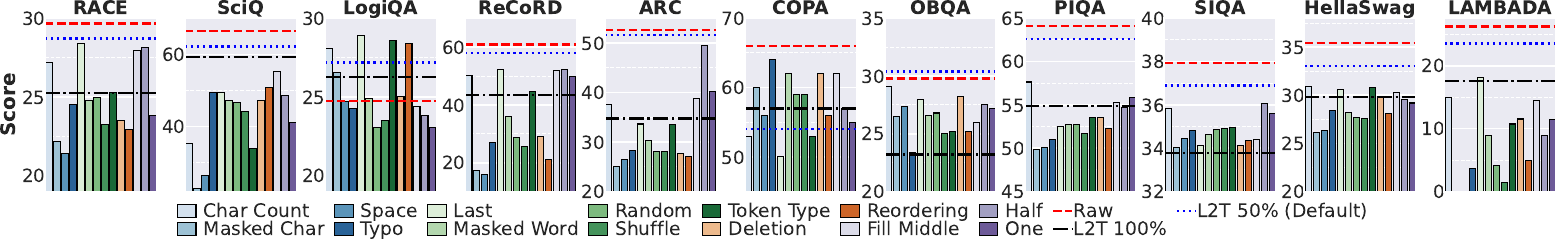}
\caption{
General benchmark performance comparison between different L2T models trained on specific 25B token single task data.
}
\label{fig:single_downstream}
\end{figure*}

\begin{table}[ht]
\centering
\small
\resizebox{\linewidth}{!}{
\begin{tabular}{@{}l@{\hspace{0.5em}}l@{\hspace{0.5em}}lcccccc}
\toprule
\multicolumn{3}{c}{\textbf{Data}} & \textbf{BoolQ} & \textbf{CB} & \textbf{MultiRC} & \textbf{RTE} & \textbf{WiC} & \textbf{WSC} \\ \midrule
\multirow{4}{*}{\rotatebox{90}{500M}} & \multirow{2}{*}{\rotatebox{90}{Disj.}} & Raw & 53.1 & 39.7 & 48.6 & 56.0 & 49.7 & 38.5 \\
     & & L2T & 40.6 & 31.0 & 53.3 & 50.2 & 47.5 & 46.2 \\
     \cmidrule{2-9}
     & \multirow{2}{*}{\rotatebox{90}{Shar.}} & Raw & 56.5 & 27.3 & 53.9 & 48.7 & 50.0 & 41.3 \\
     & & L2T & 59.1 & 31.2 & 51.1 & 51.3 & 47.3 & 51.9 \\ \midrule
\multirow{4}{*}{\rotatebox{90}{1B}} & \multirow{2}{*}{\rotatebox{90}{Disj.}} & Raw & 55.7 & 21.3 & 49.4 & 52.0 & 50.5 & 44.2 \\
     & & L2T & 48.4 & 33.6 & 51.3 & 56.0 & 48.6 & 45.2 \\
     \cmidrule{2-9}
     & \multirow{2}{*}{\rotatebox{90}{Shar.}} & Raw & 40.7 & 33.8 & 45.0 & 48.7 & 50.9 & 59.6 \\
     & & L2T & 58.0 & 25.2 & 49.4 & 49.5 & 51.3 & 42.3 \\
     \midrule
\end{tabular}
}
\caption{Evaluation results on SuperGLUE tasks for models at the 500M and 1B parameter scales. Metrics reported are accuracy (Acc) for all tasks except CB, which uses the F1 score.}
\label{tab:superglue_results}
\end{table}

\section{Analysis on Additional General Benchmarks}
\label{appendix:analysis-superglue}

We conduct an additional evaluation on SuperGLUE~\citep{NEURIPS2019_4496bf24} to analyze the broader impact of the L2T framework on Natural Language Understanding.
While linguistic competence on BLiMP consistently improves (\S\ref{subsec:results-linguistic}), results on SuperGLUE exhibit variability across different data scenarios.
This variability likely originates from the tension between structural induction and factual reinforcement~\citep{FedorenkoVarley2016LanguageThought} as discussed in \S\ref{subsec:results-general}.
Knowledge-intensive tasks, such as BoolQ and CB, depend on the factual repetition inherent in standard causal language modeling.
Consequently, performance decreases in the Disjoint scenario because the exposure to raw text is reduced.
Conversely, L2T benefits tasks that require structural resolution, such as MultiRC, by enhancing sensitivity to syntactic dependencies beyond simple word probabilities.
The size of the model also modulates these effects.
The limited capacity of models with 500M or 1B parameters likely results in high sensitivity to the ratio of raw text necessary for general reasoning.
Ultimately, these results indicate that L2T serves as a targeted linguistic force multiplier.
The framework achieves higher effectiveness when balanced with sufficient raw text to anchor general factual reasoning.

\section{Efficacy of Individual Tasks} \label{appendix:analysis-task}

A closer examination of individual linguistic phenomena (Figure \ref{fig:single_blimp}) reveals that while almost all tasks (excluding \textit{Space}) outperform the Raw model on Island effects, none succeed on Fill Gap. While the structural scaffolding from the tasks across varying granularity assists in detecting island violations, the complexity of Fill Gap requires targeted signals to capture moved elements and long-distance hierarchical dependencies. Current L2T data likely lacks the specific coverage needed for these complex structures.

\begin{figure*}[!t]
\centering
\includegraphics[
    trim={0 0.18cm 0 0}, clip,
    width=0.85\textwidth
]{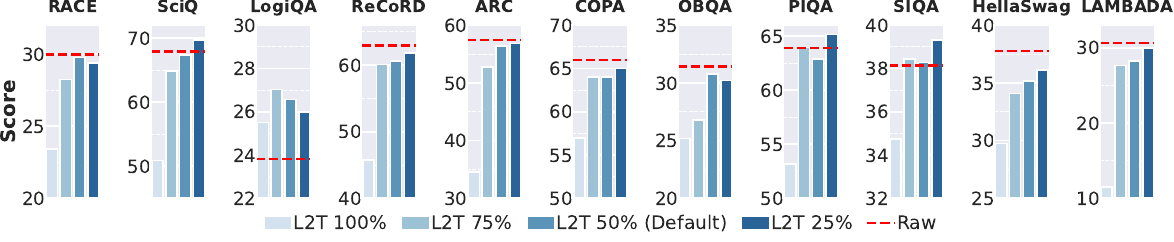}
\caption{
Performance on general benchmarks for 500M models pre-trained with different mixing ratios of standard Raw vs. L2T data for 100B tokens. L2T 100\% stands for no standard raw text mixed, i.e. 100\% L2T data.
}
\label{fig:mix_downstream}
\end{figure*}

\begin{figure}[!t]
\centering
\includegraphics[width=0.8\columnwidth]{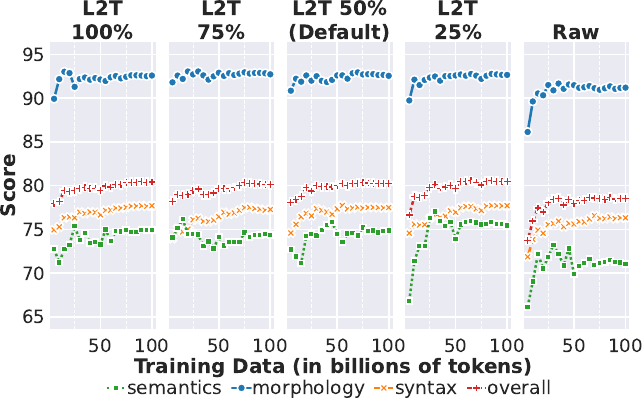}
\caption{
Linguistic competence comparisons by linguistic subfield on BLiMP between Raw and L2T 500M models with different mixing ratios of standard raw text.
100\% stands for no standard raw text mixed.
}
\label{fig:blimp_mix_subfield}
\end{figure}

\section{Mixing Ratio of Raw and L2T Data} \label{appendix:analysis-mix}

We investigate the influence of the proportion of raw text relative to L2T data by varying the L2T mixing ratio at 100\% (denoting zero raw text), 75\%, 50\% (the default), and 25\% for 500M models using 100B tokens in the Disjoint setting.\footnote{Due to computational limits, experiments use only the 500M model.}
Here, a 100\% mixing ratio denotes training solely on L2T data.

\paragraph{Linguistic Competence.}
Results on BLiMP (Figure \ref{fig:blimp_mix_subfield}) show that after training, all L2T models perform similarly across linguistic subfields, regardless of the mixing ratio.
Differences remain minor, with maximum deltas of 1.1 in semantics, 0.17 in morphology, and 0.73 in syntax.
However, early in training (e.g., 5B tokens), performance gains frequently correlate with increased L2T data, except for the L2T 100\% setting.
For instance, at 5B tokens, L2T 75\% outperforms L2T 50\% and L2T 25\% in morphology (91.8 vs. 90.8 and 89.7) and semantics (74.1 vs. 72.7 and 66.8).
This suggests that while the mixing ratio influences the initial learning trajectory, L2T data stimulates linguistic competence regardless of the specific ratio by the end of training.

\paragraph{General Benchmarks.}
While linguistic gains are stable, evaluation on general benchmarks (Figure \ref{fig:mix_downstream}) underscores the necessity of the raw text proportion (i.e., allocating sufficient training steps to raw text).
The L2T 100\% configuration, which contains no raw text, exhibits substantial performance drops, such as a 23-point decline on ARC, compared to the Raw model.
Increasing the proportion of raw text mitigates this gap; for example, the L2T 75\% setting (i.e., containing 25\% raw text) differs by only 4.7 points on ARC.
Performance generally improves as the proportion of raw text increases (except on LogiQA), a trend supported by Kendall tau correlations~\cite{665905b2-6123-3642-832e-05dbc1f48979} ranging from 0.67 to 1.0.
These results demonstrate that while L2T data enhances linguistic competence, raw text remains essential for broad knowledge and reasoning.
This aligns with the view of \citet{cheng-etal-2024-instruction} that mixing raw text is vital for retaining broad world knowledge.
Consequently, achieving an appropriate allocation balance between these data types is imperative.
Even 25\% of L2T data substantially improves linguistic competence (e.g., 1.9 overall gain over Raw in BLiMP), while at least 25\% of raw text is essential for robust general capabilities (e.g., an 18-point gain on ARC compared to L2T 100\%).

\section{Qualitative Analysis} \label{appendix:qualitative}
We conduct a qualitative analysis to examine the behavior and limitations of models trained on L2T.
First, we observe a substantial improvement of 23.2 on the ``coordinate structure constraint complex left branch'' task within island effects (Island), specifically in the 500M Disjoint setup.
This task requires distinguishing between minimally different sentences such as:\looseness=-1
\begin{quote}
\footnotesize
\setlength{\leftskip}{-1.5em}
\setlength{\rightskip}{-2em}
\textbf{Correct:} Whose \textit{mice} can Julia bring and Brett notice? \\
\textbf{Incorrect:} Whose can Julia bring \textit{mice} and Brett notice?
\end{quote}
\noindent The challenge lies in detecting subtle syntactic violations caused by the misplacement of constituents.
We attribute improvements on such tasks largely to the exposure of the model to diverse structural objectives within L2T, which collectively enhance sensitivity to complex syntactic dependencies beyond simple word probabilities.

\begin{table*}[tbh]
\centering
\small
\begin{tabularx}{\textwidth}{l l X X}
\toprule
\textbf{Category} & \textbf{Dataset} & \textbf{Source Link} & \textbf{License} \\
\midrule
\textbf{RC} & RACE~\cite{lai-etal-2017-race} & \url{http://www.cs.cmu.edu/~glai1/data/race/} & Custom (Research) \\
& SciQ~\cite{welbl-etal-2017-crowdsourcing} & \url{https://allenai.org/data/sciq} & CC BY-NC 3.0 \\
 & LogiQA~\cite{10.5555/3491440.3491941} & \url{https://github.com/lgw863/LogiQA-dataset} & No license found \\
 \midrule
 \textbf{RC+CR} & ReCoRD~\cite{zhang2018recordbridginggaphuman} & \url{https://sheng-z.github.io/ReCoRD-explorer/} & Apache 2.0 + \href{https://archive.org/about/terms}{Internet Archive's Terms of Use} \\
\midrule
\textbf{CR} & ARC (Easy)~\cite{clark2018thinksolvedquestionanswering} & \url{https://allenai.org/data/arc} & CC BY-SA 4.0 \\
 & COPA~\cite{gordon-etal-2012-semeval} & \url{https://people.ict.usc.edu/~gordon/copa.html} & No license found \\
 & OpenBookQA~\cite{mihaylov-etal-2018-suit} & \url{https://allenai.org/data/open-book-qa} & Apache 2.0 \\
 & Social IQa~\cite{sap-etal-2019-social} & \url{https://huggingface.co/datasets/allenai/social_i_qa} & CC BY 4.0 \\
 & HellaSwag~\cite{zellers-etal-2019-hellaswag} & \url{https://rowanzellers.com/hellaswag/} & MIT \\
\midrule
\textbf{Language Modeling} & LAMBADA~\cite{paperno-etal-2016-lambada} & \url{https://zenodo.org/records/2630551} & CC BY 4.0 \\
\bottomrule
\end{tabularx}
\caption{Summary of Datasets, Sources, and Licenses}
\label{tab:dataset_summary}
\end{table*}

In contrast, we observe a 7.9 point drop on the ``wh vs that with gap long distance'' task within filler gap (Fill Gap), which tests whether a long-distance dependency is correctly licensed, as shown below:
\begin{quote}
\setlength{\leftskip}{-1.5em}
\setlength{\rightskip}{-1em}
\footnotesize
Phillip forgot [\textbf{Correct:} \textit{what}] [\textbf{Incorrect:} \textit{that}] some senator that was escaping from Stacy goes to.
\end{quote}
\noindent This phenomenon hinges on tracking hierarchical syntactic relationships across intervening clauses.
The drop in performance suggests that while L2T effectively captures structural constraints, it continues to struggle with dependencies spanning extensive contexts.
This remains an area where even humans perform modestly (75\% accuracy) and LMs have often faced challenges~\cite{da-costa-chaves-2020-assessing,warstadt-etal-2020-blimp-benchmark}.\looseness=-1

\section{License} \label{appendix:license}

This study uses publicly available datasets with different licenses, as detailed in Table \ref{tab:dataset_summary}.
We also use a tokenizer file of Mistral available at \url{mistralai/Mistral-7B-v0.1}, licensed under Apache 2.0.
Note that all permit their use for academic research.

\section{Use of Generative AI Tools} \label{appendix:genai}
The authors acknowledge the use of LLMs during the preparation of this work. Gemini 3.0 Pro were utilized to find related work and to improve the grammar and clarity of the draft. Additionally, GPT-5 served as a coding assistant for implementation and debugging.

\end{document}